\documentclass[11pt,a4paper]{article}
\usepackage{times,latexsym}
\usepackage{url}
\usepackage[T1]{fontenc}
\usepackage{mdframed}
\usepackage{multirow}
\usepackage{makecell}
\usepackage{graphicx}

\usepackage{booktabs}
\usepackage{enumitem}

\usepackage[acceptedWithA]{tacl2021v1}
\usepackage{xspace,mfirstuc,tabulary}

\newif\iftaclinstructions
\taclinstructionsfalse % AUTHORS: do NOT set this to true
\iftaclinstructions
\renewcommand{\confidential}{}
\renewcommand{\anonsubtext}{(No author info supplied here, for consistency with
TACL-submission anonymization requirements)}
\newcommand{\instr}
\fi

\iftaclpubformat % this "if" is set by the choice of options

\else

\fi

\title{Benchmarking Large Language Models for Conversational Question Answering in Multi-instructional Documents}

\author{
  Shiwei Wu$^1$,
  Chen Zhang$^2$,
  Yan Gao$^2$,
  Qimeng Wang$^2$,
  Tong Xu$^1$,
  Yao Hu$^2$,
  Enhong Chen$^1$,
  \\
  \ \\
  $^1$University of Science and Technology of China
  $^2$Xiaohongshu.Inc
  \\
  \texttt{dwustc@mail.ustc.edu.cn, \{zhangchen3, yadun, qimengwang,} 
  \\
  \texttt{xiahou\}@xiaohongshu.com, \{tongxu, cheneh\}@ustc.edu.cn}
}

\date{}

\begin{document}
\maketitle
\begin{abstract}
    Instructional documents are rich sources of knowledge for completing various tasks, yet their unique challenges in conversational question answering (CQA) have not been thoroughly explored. Existing benchmarks have primarily focused on basic factual question-answering from single narrative documents, making them inadequate for assessing a model’s ability to comprehend complex real-world instructional documents and provide accurate step-by-step guidance in daily life.    
    To bridge this gap, we present \texttt{InsCoQA}, a novel benchmark tailored for evaluating large language models (LLMs) in the context of CQA with instructional documents. Sourced from extensive, encyclopedia-style instructional content, \texttt{InsCoQA} assesses models on their ability to retrieve, interpret, and accurately summarize procedural guidance from multiple documents, reflecting the intricate and multi-faceted nature of real-world instructional tasks.
    Additionally, to comprehensively assess state-of-the-art LLMs on the \texttt{InsCoQA} benchmark, we propose \textsc{InsEval}, an LLM-assisted evaluator that measures the integrity and accuracy of generated responses and procedural instructions.
\end{abstract}
\section{Introduction}

\begin{figure}[t]
\centering
\hspace*{-0.8cm}
\includegraphics[width=1.2\linewidth, height=1.3\linewidth]{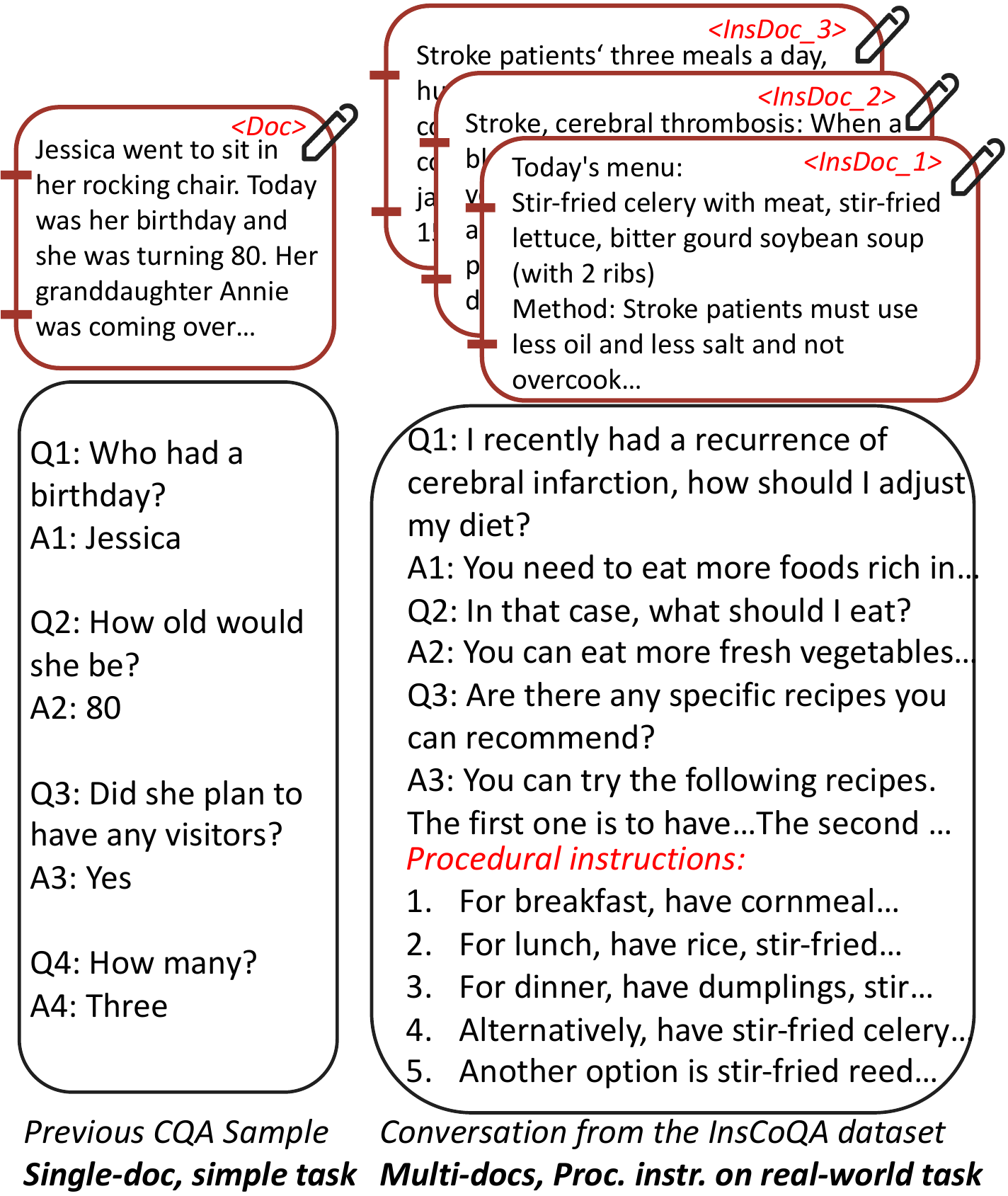}
\caption{\textbf{Comparison of Conversational Question Answering (CQA) Samples.} Our \texttt{InsCoQA} dataset provides \textit{more complex procedural guidance} derived from \textit{multiple instructional documents}, addressing \textit{intricate, real-world tasks}. In contrast, the previous CQA sample involves basic narrative elements and factual recall, focusing on simple, single-document information.}
\label{figure:teaser}
\vspace{-2mm}
\end{figure}

Instructional documents are indispensable resources for acquiring knowledge and completing a myriad of tasks, from assembling furniture to troubleshooting electronic devices. Nowadays, users on social platforms like \textit{Xiaohongshu}\footnote{www.xiaohongshu.com} and \textit{Lemon8}\footnote{www.lemon8-app.com} frequently share their experiences and gained knowledge with these tasks in everyday life. These documents provide step-by-step guidance and detailed explanations that are essential in both professional and personal contexts, enabling users to perform tasks efficiently and accurately, regardless of their prior experience.

However, effectively searching, comprehending, and summarizing helpful procedures from instructional documents presents unique challenges, especially in the realm of conversational question answering (CQA)~\cite{coqa, quac, opencoqa, squad, squadrun}. 
As illustrated in Figure~\ref{figure:teaser}, unlike previous CQA datasets~\cite{coqa, squad, quac, opencoqa}, which typically retrieve straightforward information from single-document sources such as news articles and children's stories, benchmarks on instructional documents require understanding and synthesizing information from multiple sources to provide accurate, comprehensive, and step-by-step guidance, while also generating natural responses in a conversational context.

To address this need, we introduce \texttt{InsCoQA} (\texttt{Ins}tructional Documents based \texttt{Co}nversational \texttt{Q}uestion \texttt{A}nswering), a novel benchmark specifically designed to evaluate LLMs in the context of conversational question answering (CQA) with instructional documents. Our benchmark is sourced from the \textit{Xiaohongshu} platform, which offers a rich and diverse collection of user-generated content spanning a wide range of instructional domains. Each entry in our dataset has been meticulously curated and verified by human annotators to ensure high-quality annotations that capture the intricate and multi-faceted nature of real-world instructional content.
To comprehensively evaluate model performance on our \texttt{InsCoQA} benchmark, we introduce \textsc{InsEval} (\textsc{Ins}tructional \textsc{Eval}uator), which assesses models across \textit{semantic, lexical, and quantitative} dimensions. We also conduct extensive experiments with state-of-the-art Large Language Models (LLMs) on our benchmark.
The main contributions of our work are summarized as follows:

\begin{itemize}[leftmargin=*] 
\item We propose \texttt{InsCoQA}, a novel benchmark that assesses LLMs on their ability to retrieve, interpret, and accurately summarize procedural guidance from multiple documents. \texttt{InsCoQA} is more challenging and applicable to real-world tasks compared to previous CQA benchmarks.

\item We collect 13.9k instructional conversations from \textit{Xiaohongshu} across 13 diverse domains. The dataset is meticulously curated and verified by human annotators to ensure high quality.

\item We introduce \textsc{InsEval}, a comprehensive LLM-assisted evaluator for assessing popular LLMs on the \texttt{InsCoQA} benchmark, focusing on the integrity and accuracy of procedural instructions provided for completing user-specified tasks.
% across \textit{semantic}, \textit{lexical}, and \textit{quantitative} dimensions. 
\end{itemize}
\section{Related Work}
\textbf{Conversational Question Answering (CQA)} has emerged as a critical area of research, particularly with the advent of large language models (LLMs) capable of engaging in human-like dialogue. Early benchmarks, such as CoQA~\cite{coqa} and QuAC~\cite{quac}, pioneered this task by focusing on generating coherent and contextually appropriate responses to questions based on narrative texts like news articles and story excerpts. These datasets were instrumental in advancing the field, spurring research in various areas such as question rewriting~\cite{question_rewrite, unpack_rewrite, few_rewrite}, open-domain CQA~\cite{openre_coqa, opencoqa}, multi-modal CQA~\cite{mmcoqa}, knowledge-based conversational QA~\cite{kgcoqa}, and simulating conversations with LLMs~\cite{llmtalk}.

However, previous CQA benchmarks primarily focused on basic narrative elements and factual recall, relying on simple, single-document information. With the advent of advanced LLMs~\cite{chatgpt, gpt4, gpt4o, llama1, llama2, llama3, mistral}, these benchmarks have become relatively easy to tackle. In contrast, our proposed \texttt{InsCoQA} benchmark addresses the unique challenges of CQA with instructional documents, requiring models to synthesize procedural guidance from multiple sources and provide complex, step-by-step instructions.
\section{Constructing Conversational QA Dataset for Multiple Instructional Documents}

\begin{figure*}[!t]
\centering
\includegraphics[width=0.95\linewidth]{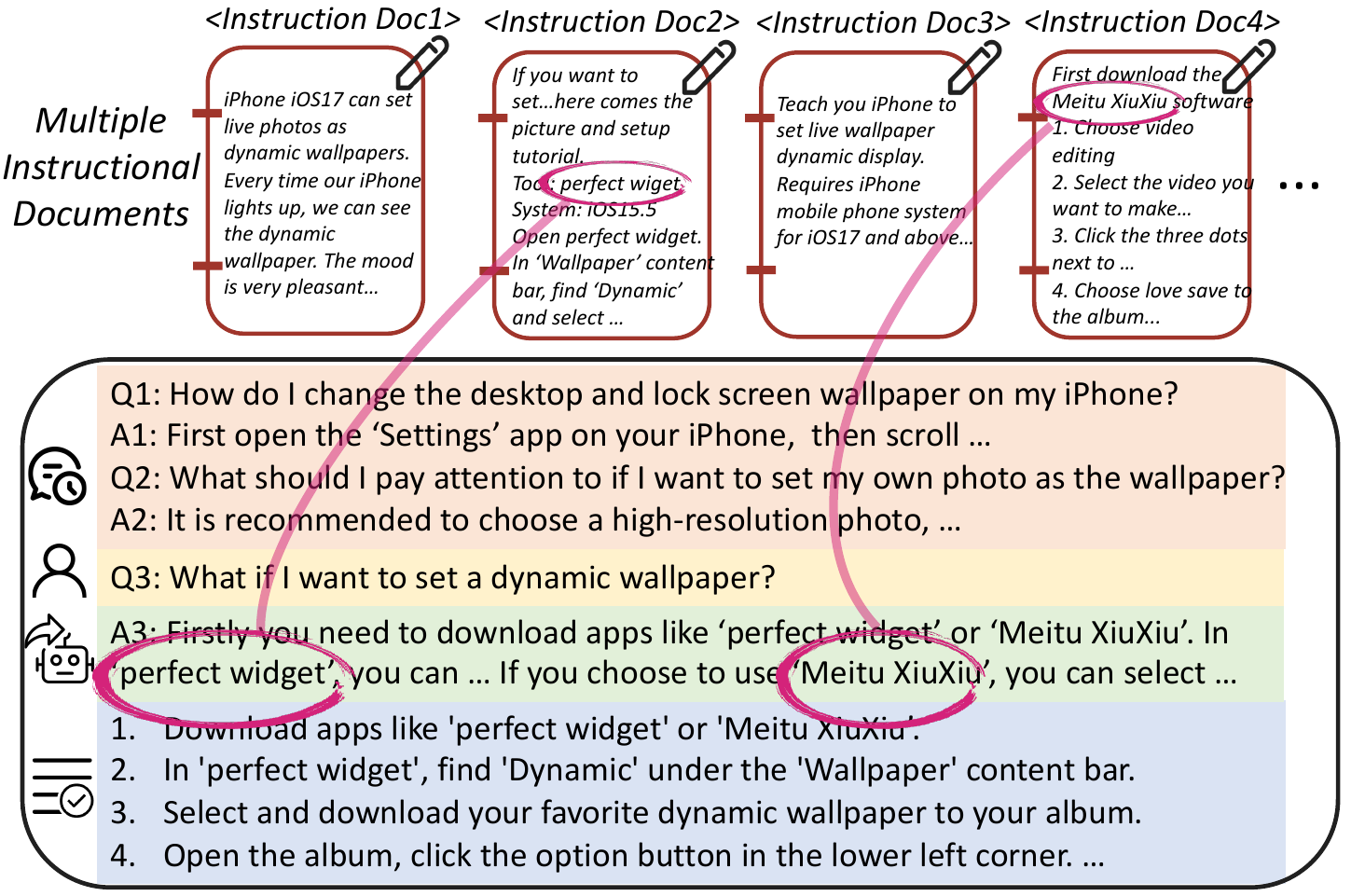}
\caption{An instructional example from the \texttt{InsCoQA} dataset, where the conversational QA is closely aligned with the referenced instructional documents. This sample highlights key features such as multi-document references and a focus on providing procedural instructions for real-world tasks.}
\label{figure:illustration}
\end{figure*}

\begin{figure*}[!t]
\centering
\includegraphics[width=0.95\linewidth]{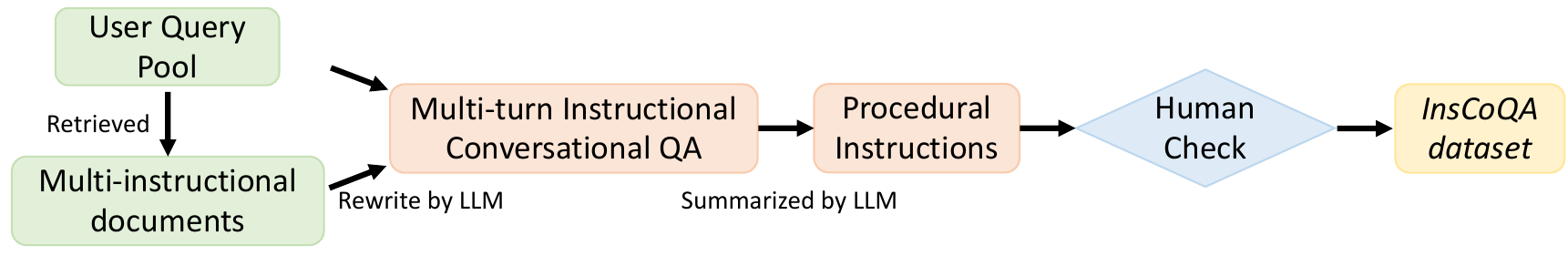}
\caption{\textbf{Pipeline for constructing the \texttt{InsCoQA} dataset.} We begin by retrieving multiple instructional documents relevant to the instructional-intended user query, then rewriting them into multi-turn instructional conversational Q\&A. The responses are further summarized into procedural instructions. The \texttt{InsCoQA} dataset is finally constructed after a thorough human check.}
\label{figure:pipeline}
\end{figure*}

To enhance the document understanding and task-completion capabilities of large language models (LLMs), we collected massive instructional documents from the \textit{Xiaohongshu} platform, and further constructed a Conversational QA dataset for multi-instructional documents, termed the \texttt{InsCoQA} dataset. The data collection and processing approaches are detailed below, with the overall pipeline illustrated in Figure~\ref{figure:pipeline}.

\subsection{Collecting Instructional Documents for Popular User Queries}
To gather high-quality, instruction-focused queries, we extracted the most frequently asked user queries from the \textit{Xiaohongshu} platform. Non-instructional, incomplete, semantically ambiguous, subject-specific, and time-sensitive queries were filtered out using OpenAI GPT-4~\cite{gpt4}. The detailed prompt for this filtering process is provided in Appendix~\ref{supp:query_filter_prompt}.

Further, for each filtered high-quality query, we retrieve multiple relevant documents from the \textit{Xiaohongshu} platform using the internal information retrieval engine~\cite{pagerank,bert,roberta}. Notably, the retrieved documents are typically instructional, providing comprehensive guidance and knowledge to address the related user queries, as shown in Figure~\ref{figure:teaser} and \ref{figure:illustration}. 
To emulate the human behavior of referencing multiple sources when responding to questions, we collect multiple documents for each conversational QA sample. Different from \cite{coqa, quac} which typically uses a single document as the reference, our task requires LLMs to identify important commonalities among different documents and provide comprehensive procedural instructions to complete the task.

\subsection{Formatting to Conversational QA Dialog}
It is essential for AI systems to understand and assist humans in conversations with interconnected questions and answers. Therefore, we format the task as a multi-turn conversational dialog, where the questions are conversational and the answers are free-form text. As shown in Figure~\ref{figure:illustration}, the conversational questions exhibit challenging phenomena such as coreference and pragmatic reasoning, which require LLMs to understand the conversation context.

We construct coherent conversations to assess the LLM’s ability to retrieve external knowledge and understand contextual information. Given a selected user query and the corresponding multiple documents, we use GPT-4~\cite{gpt4} to generate multi-turn conversational history and the final-turn question. Subsequently, responses are generated and summarized into procedural instructions based on the user query, multiple reference documents, and the conversation history.
The prompt for generating the conversational QAs and procedural instructions is detailed in Appendix~\ref{supp:history_generation_prompt} and \ref{supp:response_generation_prompt}.

To improve the quality of our \texttt{InsCoQA} dataset, we recruit three human annotators to filter out issues such as (1) reversal of question and answer, (2) common hallucinations in responses, (3) marketing information, and (4) security concerns including politics, ethics, and NSFW content. Additionally, we inspect 10\% of the dataset to ensure quality, achieving an accuracy rate of 95.15\%.
Furthermore, we translate the entire dataset into English using GPT-4~\cite{gpt4}.
\section{Dataset analysis}

In this section, we provide a detailed analysis of the \texttt{InsCoQA} dataset as follows.

\subsection{Data Composition}

\begin{figure}[h]
  \includegraphics[width=0.5\textwidth, height=0.25\textwidth]{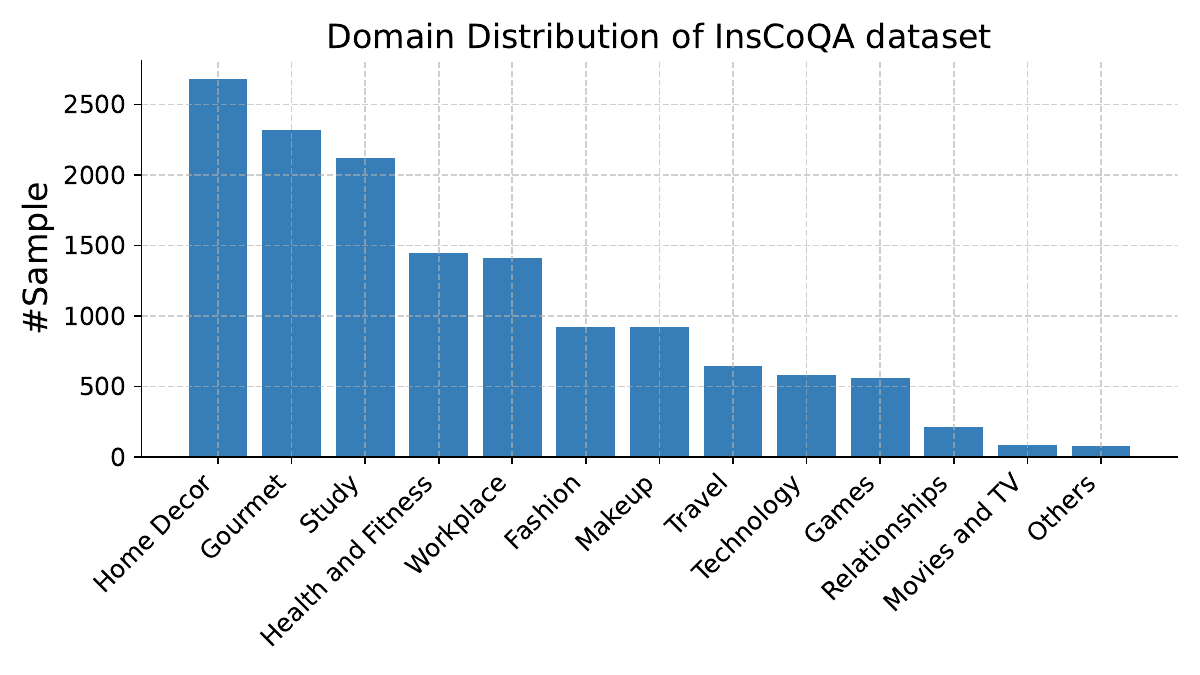}
  \caption{Domain distribution of \texttt{InsCoQA} dataset.}
  \label{fig:domain_distribution}
\end{figure}

The \texttt{InsCoQA} dataset consists of 13,959 samples, divided into 9,760 for training, 1,406 for validation, and 2,793 for testing. The dataset is categorized into 13 different domains, chosen to ensure diversity in the tasks covered and complexity in the instructions provided as shown in Figure~\ref{fig:domain_distribution}.

\subsection{Data Statistics}
\begin{table*}[t]
\centering
\scriptsize
\begin{tabular}{c|ccc|ccc|ccc|ccc|ccc}
\hline
& \multicolumn{3}{c|}{\#Instruct Documents} & \multicolumn{3}{c|}{Instruct Doc. \#Word} & \multicolumn{3}{c|}{Instruct Doc. \#Token} & \multicolumn{3}{c|}{Procedure \#Word} & \multicolumn{3}{c}{Procedure \#Step} \\
\hline
& Min & Max & Mean & Min & Max & Mean & Min & Max & Mean & Min & Max & Mean & Min & Max & Mean \\
\hline
Training & 1 & 44 & 4.07 & 15 & 1024 & 144 & 23 & 1478 & 204 & 5 & 292 & 82 & 1 & 36 & 6.5 \\
\hline
Validation & 1 & 14 & 4.18 & 12 & 699 & 138 & 22 & 986 & 196 & 10 & 272 & 84 & 2 & 28 & 6.5 \\
\hline
Test & 1 & 8 & 4.10 & 10 & 770 & 141 & 18 & 1032 & 200 & 10 & 404 & 83 & 1 & 23 & 6.6 \\
\hline
\end{tabular}
\caption{Statistics of referenced documents and procedural instructions in \texttt{InsCoQA} dataset.}
\label{tab:statistics}
\end{table*}

We provide a detailed breakdown of the dataset statistics to emphasize the complexity and diversity of the \texttt{InsCoQA} dataset, as illustrated in Table~\ref{tab:statistics}. On average, each conversation consists of 3.11 Q/A rounds, with each conversation referencing approximately 4.09 instructional documents with 606 words. For the final round question in each conversation, an average of 6.50 summarized instructional procedures are provided. The distribution of instructional documents and procedures is well-balanced across various domains as shown in Table~\ref{tab:domain_statistics}, reflecting the broad applicability of \texttt{InsCoQA} dataset in capturing real-world instructional tasks.

\begin{table}
\footnotesize
\hspace{-1em}
\setlength{\tabcolsep}{0.1mm} % Adjust this length to change the column spacing
\begin{tabular}{p{2.11cm}rccc}
\toprule
Domain & \#Instr. Doc. & \#Instr. Doc. & \#Q/A & \#Instructional \\
 & /conversation & length & rounds & procedures \\
\midrule
Fashion  & 4.36 & 558 & 3.09 & 5.67 \\
Study  & 3.98 & 570 & 3.22 & 6.28 \\
Gourmet & 4.36 & 630 & 3.04 & 7.17 \\
Makeup & 4.28 & 613 & 3.09 & 6.36 \\
Movies \& TV & 4.15 & 537 & 3.20 & 6.16 \\
Workplace & 3.71 & 609 & 3.17 & 6.75 \\
Relationships & 4.43 & 717 & 3.06 & 6.52 \\
Home Decor & 4.06 & 600 & 3.16 & 6.39 \\
Games & 3.69 & 509 & 3.08 & 6.44 \\
Travel & 3.77 & 636 & 3.01 & 5.91 \\
Health \& Fitness & 4.22 & 685 & 3.02 & 6.32 \\
Technology & 4.02 & 558 & 3.13  & 7.24 \\
Others & 4.13 & 643 & 3.22  & 7.06 \\
\midrule
Total average & 4.09 & 606 & 3.11 & 6.50 \\
\bottomrule
\end{tabular}
% \caption{Distribution of domains in \texttt{InsCoQA} dataset.}
\caption{Average statistics across domains in the \texttt{InsCoQA} dataset. The distribution of instructional documents and procedures is well-balanced across various domains.}
\label{tab:domain_statistics}
\end{table}

\subsection{Data Characteristics}
As illustrated in Figure~\ref{figure:teaser} and Figure~\ref{figure:illustration}, the \texttt{InsCoQA} dataset is different from the previous datasets~\cite{coqa,quac} in the following three main aspects:
\begin{itemize}[leftmargin=*]
    \item \textbf{Multi-documents Information Gathering}: Unlike traditional CQA datasets that primarily focus on retrieving information from single documents, \texttt{InsCoQA} challenges models to gather, synthesize, and reason over information spread across multiple instructional documents. This requires a deeper level of comprehension and the ability to interrelate procedural steps that may be dispersed across different sources.
    % Unlike previous CQA datasets that focus on single-document retrieval, \texttt{InsCoQA} requires models to synthesize information from multiple instructional documents to provide accurate and comprehensive answers.
    \item \textbf{Procedural Instructions Summarization}: In contrast to merely retrieving factual information, \texttt{InsCoQA} emphasizes the need for models to summarize and articulate step-by-step procedural guidance. This involves not only extracting relevant details but also organizing them logically and coherently to guide the user through a task effectively. 
    \item \textbf{Focus on Complex, Real-world Tasks}: The tasks within \texttt{InsCoQA} are derived from real-world instructional content, reflecting the intricacy and nuanced nature of actual user experiences. The dataset includes scenarios where the steps are non-trivial and require detailed understanding, such as multi-stage assembly processes or troubleshooting guides that involve conditional steps based on user input.
\end{itemize}
\section{\textsc{InsEval}: Comprehensive Evaluation for \texttt{InsCoQA} benchmark}
To assess LLM’s ability to solve tasks and provide accurate instructions, we ask the models to generate \textit{Plain-text response} and then summarize to \textit{Procedural instructions} for each conversational question. Given the free-form nature of responses and the open-set nature of instruction candidates, traditional reference-based recall metrics~\cite{coin, assistq} often fall short in capturing the integrity and accuracy of the answer.

To address these limitations, and drawing inspiration from recent works~\cite{llava, gpteval, mm-narrator}, we introduce the \textsc{Ins}tructional \textsc{Eval}uator (\textsc{InsEval}). \textsc{InsEval} first evaluates the quality of both the \textit{Plain-text response} and the \textit{Procedural instructions} using an LLM-assisted approach. It then quantifies the alignment between the \textit{Plain-text response} and the ground truth by employing both character-level and word-level ROUGE-L metrics. The details of \textsc{InsEval} are outlined below.

\subsection{LLM-assisted Evaluation Metrics}
Evaluating whether the generated instructions accurately address the question and cover the correct procedures is challenging and often inaccurate when relying solely on exact text matching with the ground truth. To improve the evaluation process, we utilize an additional LLM, specifically Qwen2-7B-Instruct~\cite{qwen2} by default, to assess both the \textit{Plain-text response} and the \textit{Procedural instructions} using two defined metrics.

Given the conversational context and question, we evaluate the generated \textit{Plain-text response} and $M$ predicted instructions with an additional LLM acting as an expert evaluator to calculate the following metrics:

\begin{itemize}[leftmargin=*]
\item {\bf Judge Score:} This metric assesses how well the \textit{Plain-text response} addresses the user’s concerns, considering both the question and the conversation history. The response is compared to the ground truth, and a score between 1 and 10 is assigned based on the helpfulness and relevance of the answer.

\item {\bf Task Completion Rate:} This metric determines whether the subset $M_c \subseteq M$ of predicted \textit{Procedural instructions} properly summarizes the correct procedures and answers the question. The recall of correctly answered instructions is defined as the \textit{Task Completion Rate}, calculated as follows:

$$
Task\ Completion\ Rate = \frac{|M_c|}{K}
$$

\end{itemize}

The prompt used for the above LLM-assisted metrics is detailed in Appendix~\ref{supp:judge_prompt}.

\begin{table*}[h]
% \vspace{-3mm}
\setlength{\tabcolsep}{1pt}
\small
\centering
\begin{tabular}{l|c|cccc}
\toprule
\multirow{2}{*}{Method} & \multirow{2}{*}{Parameters} & \multicolumn{4}{c}{Performance on \texttt{InsCoQA} dataset} \\
& & \textit{JudgeScore} & \textit{TaskCompletionRate} & \textit{W-ROUGE-L} &  \textit{C-ROUGE-L} \\
\midrule
\multicolumn{5}{l}{\textbf{Validation set}} \\
\midrule
% gemma-2 & 27B & 2.43 & 2.04 & 45.1\% & 48.1\% \\
Mistral-Instruct-v0.3~\cite{mistral} & 7B & 7.45 & 78.83\% & 0.24 & 0.47 \\
WizardLM-2~\cite{wizardlm2} & 7B & 7.89 & 85.64\% & 0.18 & 0.39 \\
Llama-3.1-Instruct~\cite{llama3} & 8B & 7.66 & 79.38\% & 0.25 & 0.48 \\

GPT-4~\cite{gpt4} & - & 8.45 & 90.77\% & 0.28 & 0.50 \\
\midrule
\multicolumn{5}{l}{\textbf{Test set}} \\
\midrule
Mistral-Instruct-v0.3~\cite{mistral} & 7B & 7.46 & 83.38\% & 0.26 & 0.48 \\
WizardLM-2~\cite{wizardlm2} & 7B & 7.85 & 87.83\% & 0.19 & 0.39 \\
Llama-3.1-Instruct~\cite{llama3} & 8B & 7.49 & 82.98\% & 0.26 & 0.49 \\
GPT-4~\cite{gpt4} & - & 8.55 & 91.37\% & 0.31 & 0.52 \\
\bottomrule
\end{tabular}
\caption{Performance of various LLMs on the \texttt{InsCoQA} dataset, evaluated across four metrics used by \textsc{InsEval}.}
% \vspace{-3mm}
\label{tab:exp}
\end{table*}

\subsection{Text-matching Evaluation Metrics}
\textsc{InsEval} further quantifies the alignment between the \textit{Plain-text response} and the ground truth using both character-level and word-level ROUGE-L metrics.

\begin{itemize}[leftmargin=*]
\item {\bf ROUGE-L:} This metric~\cite{rouge} operates on the principle that the longer the Longest Common Subsequence (LCS) between two sentences, the greater their similarity. The LCS-based F-measure is used to estimate the similarity between two sentences $X$ of length $m$ and $Y$ of length $n$, as defined below:

$$
R_{lcs} = \frac{LCS(X,Y)}{m}
$$

$$
P_{lcs} = \frac{LCS(X,Y)}{n}
$$

$$
F_{lcs} = \frac{(1+\beta^2)R_{lcs}P_{lcs}}{R_{lcs}+\beta^2P_{lcs}}
$$

Specifically, word-level ROUGE-L focuses on the accuracy of specific words by calculating the $F_{lcs}$ for words in the sentences, while character-level ROUGE-L takes into account word forms, grammar, and punctuation by calculating the $F_{lcs}$ for characters. The parameter $\beta$ controls the relative importance of recall ($R_{lcs}$) and precision ($P_{lcs}$) in the F-measure, and we default it to 1.

\end{itemize}

\subsection{Experimental Results on LLMs}
To evaluate the ability of current LLMs in generating accurate answers and procedural instructions, we conducted an experimental comparison of several popular LLMs, as shown in Table~\ref{tab:exp}.

We concatenate the referenced documents, conversation history, and the latest question to form the input, and prompt the LLMs to generate both a plain-text response and a summarized set of procedural instructions. The detailed prompt is provided in Appendix~\ref{supp:inference_prompt}.
Among the evaluated models, GPT-4~\cite{gpt4} demonstrated the best performance, excelling at understanding conversational questions and accurately summarizing procedural instructions. Llama-3.1-Instruct-8B~\cite{llama3} showed better task completion and instruction-following capabilities compared to Mistral-Instruct-v0.3-7B~\cite{mistral}. WizardLM-2-7B~\cite{wizardlm2} outperforms other 7/8B-level LLMs in response quality and task completion, despite lower text-alignment with the ground truth, as reflected by its higher \textit{Judge Score}/\textit{Task Completion Rate}, and lower \textit{ROUGE-L} scores.

\noindent \textbf{Ablation on different LLM evaluator.} 
We demonstrate that smaller LLMs, such as Qwen2-7B-Instruct~\cite{qwen2}, are sufficiently competent for evaluating answer quality when compared to larger models like GPT-4~\cite{gpt4}. As shown in Table~\ref{tab:inseval_ablation}, while the absolute values of the \textit{Judge Score} and \textit{Task Completion Rate} may differ due to inherent biases in the LLMs, the ranking of the evaluated LLMs remains consistent across both evaluators.

\begin{table}[h]
% \vspace{-3mm}
\setlength{\tabcolsep}{3pt}
\small
\centering
\begin{tabular}{l|cc|cc}
\toprule
\multirow{2}{*}{Method} & \multicolumn{2}{c|}{Qwen2-7B-Inst.} & \multicolumn{2}{c}{GPT-4} \\
& \textit{JS} & \textit{TCR} & \textit{JS} & \textit{TCR} \\
\midrule
Mistral-Inst.-v0.3-7B & 7.45 & 78.83\% & 6.24 & 68.83\% \\
WizardLM-2-7B & 7.89 & 85.64\% & 6.72 & 72.38\% \\
Llama-3.1-Instruct-8B & 7.66 & 79.38\% & 6.43 & 69.92\% \\
GPT-4 & 8.45 & 90.77\% & 7.12 & 79.64\% \\
% Llama-3-70B & & \\
% Qwen-2-72B & & \\
% WizardLM-2-8x22B & & \\
\bottomrule
\end{tabular}
\caption{Evaluation performance using different LLM evaluators on the validation set of \textsc{InsCoQA}. The results demonstrate that smaller LLMs are sufficiently competent for assisting in evaluating the accuracy of summarized procedures. \textit{JS} and \textit{TCR} refer to \textit{Judge Score} and \textit{Task Completion Recall}, respectively.}
% \vspace{-3mm}
\label{tab:inseval_ablation}
\end{table}
\section{Conclusion}
In this paper, we introduced \texttt{InsCoQA}, a novel benchmark tailored to evaluate the capabilities of large language models (LLMs) in conversational question answering (CQA) involving instructional documents. \texttt{InsCoQA} presents a unique challenge for models by requiring them to navigate complex, real-world tasks, extracting and summarizing procedural guidance from multiple instructional sources. 
Additionally, we proposed \textsc{InsEval}, an LLM-assisted evaluator that measures the integrity and accuracy of generated responses and procedural instructions produced by state-of-the-art LLMs on this benchmark.

\bibliography{tacl2021}
\bibliographystyle{acl_natbib}
\appendix
\clearpage
\section{Query Filtering Prompt} \label{supp:query_filter_prompt}

% \noindent \textit{Query Filtering Prompt:} 
\begin{mdframed} 
\setlength{\parindent}{0pt}
You are tasked with evaluating queries to determine if they are suitable for a conversational Question Answering task involving instructional documents. Each selected query should be:
    
1. Complete and clear.
     
2. Objective.
     
3. Have an answer that does not vary depending on the current situation.
     
4. Focused on instructional content, where the instructions typically guide the user to complete a task.
    
Please respond with “Yes/No” for each query.

\end{mdframed}

\section{Conversational QA History Generation Prompt} \label{supp:history_generation_prompt}
% \noindent \textit{Conversational QA History Generation Prompt:}
\begin{mdframed} 
\setlength{\parindent}{0pt}
You are tasked with generating a multi-turn conversation history based on a given query. Utilize the provided query to create a series of interactions that reflect a coherent and contextually rich dialogue.

The input and output should be formatted as JSON, as illustrated below:

Input: \{``query'': \texttt{[QUERY]}\}

Output: \{``history'': [\texttt{[[HISTORY\_Q0], [HISTORY\_A0]], [[HISTORY\_Q1], [HISTORY\_A1]], ...}], (Generated conversation history, segmented into multiple rounds, with each round comprising a question and a corresponding reply)

``question'': \texttt{[QUESTION]}, (the last question in the generated conversation history)

Based on the above guidelines, please generate the following output:

Input:
\{
``query'': \texttt{[QUERY]}
\}

Output:
\end{mdframed}

\section{Response and Instructional Procedures Generation Prompt} \label{supp:response_generation_prompt}
\begin{mdframed}
\setlength{\parindent}{0pt}
    You are tasked with filtering a conversational QA dataset to identify content suitable for instructional documents. You will be provided with the conversation history, the last round question, and the corresponding answer.
    
    Your goal is to determine if the final response is appropriate for instructional purposes. A qualified sample should meet the following criteria:
    
    1. The conversation history and question should be informative, avoiding trivial statements like "Hello, I have a question to ask" or "Okay, what is the question about?"
    
    2. The last question should be a genuine inquiry, not statements like "Thank you for your trouble".
    
    3. The content is appropriate for instructional purposes.
    
    4. The last round response is objective and does not contain any personal opinions or subjective judgments.
    
    5. The last round response can be transformed into clear procedural instructions.
    
    If the content meets these criteria, please summarize the final response into a set of precise procedural instructions. These instructions should:
    
    1. Provide a solution to the question,
    
    2. Follow a logical and causal order.
    
    3. Be free from repetition.
    
    4. Be strictly and clearly outlined in a list format, without using ANY sublists.
    
    Output Format:
    
    If the sample is suitable: “Summarized procedural instructions: 1. xx, 2. xx, …
    
    If the sample is not suitable: "The sample is not suitable for the task.
\end{mdframed}

\section{LLM Inference Prompt} \label{supp:inference_prompt}
\begin{mdframed}
\setlength{\parindent}{0pt}
You will be given the conversation history, multiple referenced instructional documents which may contain crucial information to the task, and the last round question.

Your first task is to answer the last round question based on conversation history and multiple referenced instructional documents.

Your second task is to summarize the procedural instructions from your answer to complete the task. These procedural instructions should meet the following criteria:

1. Provide a solution to the last round question.

2. Follow a logical and causal order.

3. Be free from repetition.

4. Be strictly and clearly outlined in a list format, without using ANY sublists.

Provide your response as follows:

Answer: (your answer)

Procedural instructions: (your instructions, the format is '1. ... 2. ...')

You MUST provide values for 'Answer:' and 'Procedural instructions:' in your answer.

Now here are the conversation history, multiple referenced instructional documents, and the last round question.

Conversation history: \{conversation\_history\}

Referenced Instructional Documents: \{documents\}
The last round question: \{question\}

Provide your response STRICTLY following the format provided above.

Response format as below.

Answer:

Procedural instructions:
\end{mdframed}

\section{Judge Prompt} \label{supp:judge_prompt}
% \noindent \textit{Judge Prompt:}
\begin{mdframed}
\setlength{\parindent}{0pt}
You will be given a conversation\_history, user\_question, system\_answer, ground\_truth\_answer, system\_procedural\_instructions, and ground\_truth\_procedural\_instructions.

Your task is to provide a 'total rating' scoring how well the system\_answer answers the user concerns expressed in the user\_question, taking into account the context provided by the conversation\_history and comparing it with the ground\_truth\_answer.

This evaluation focuses on conversational QA of instructional documents.

Give your answer on a scale of 1 to 10, where 1 means that the system\_answer is not helpful at all, and 10 means that the system\_answer completely and helpfully addresses the user\_question based on the instructional documents.

You should also provide a 'task completion rate' indicating the percentage of the task that the system\_procedural\_instructions completes compared to the ground\_truth\_procedural\_instructions,
task completion rate = (number of steps in the system\_procedural\_instructions that are also in the ground\_truth\_procedural\_instructions) / (total number of steps in the ground\_truth\_procedural\_instructions), the scale is from 0 to 100.

Here is the scale you should use to build your answer:

1: The system\_answer is terrible: completely irrelevant to the question asked, or very partial

2: The system\_answer is extremely poor: misses almost all key aspects of the question

3: The system\_answer is very poor: misses many key aspects of the question

4: The system\_answer is poor: misses several key aspects of the question

5: The system\_answer is somewhat helpful: provides limited support but misses important aspects

6: The system\_answer is somewhat helpful: provides partial support but could be significantly improved

7: The system\_answer is mostly helpful: addresses most aspects of the question but with room for improvement

8: The system\_answer is good: addresses most aspects of the question well, with minor improvements needed

9: The system\_answer is very good: addresses almost all aspects of the question very well

10: The system\_answer is excellent: relevant, direct, detailed, and addresses all the concerns raised in the question

Provide your feedback as follows:

Feedback:::
Total rating: (your rating, as a number between 1 and 10)
Task completion rate: (the percentage of the task that the system\_answer completes, as a number between 0 and 100)

You MUST provide values for 'Total rating:' and 'Task completion rate:' in your answer.

Now here are the conversation history, question, system answer, ground truth answer, system procedural instructions, and ground truth procedural instructions.

Conversation history: \{conversation\_history\}

Question: \{question\}

System Answer: \{system\_answer\}

Ground Truth Answer: \{ground\_truth\_answer\}

System Procedural Instructions: \{system\_procedural\_instructions\}

Ground Truth Procedural Instructions: \{ground\_truth\_procedural\_instructions\}

Provide your feedback:

Total rating:

Task completion rate:

\end{mdframed}

% \iftaclpubformat

% \onecolumn

% \appendix
% \section{Author/Affiliation Options as set forth by MIT Press}
% \label{sec:authorformatting}

% Option 1. Author’s address is underneath each name, centered.

% \begin{quote}\centering
%   \begin{tabular}{c}
%     \textbf{First Author} \\
%     First Affiliation \\
%     First Address 1 \\
%     First Address 2 \\
%     \texttt{first.email@example.com}
%   \end{tabular}
%   \ 
%   \begin{tabular}{c}
%     \textbf{Second Author} \\
%     Second Affiliation \\
%     Second Address 1 \\
%     Second Address 2 \\
%     \texttt{second.email@example.com}
%   \end{tabular}

%   \begin{tabular}{c}
%     \textbf{Third Author} \\
%     Third Affiliation \\
%     Third Address 1 \\
%     Third Address 2 \\
%     \texttt{third.email@example.com}
%   \end{tabular}
% \end{quote}

% Option 2. Author’s address is linked with superscript characters to its name,
% author names are grouped, centered.

% \begin{quote}\centering
%     \textbf{First Author$^\diamond$} \quad \textbf{Second Author$^\dagger$} \quad
%     \textbf{Third Author$^\ddagger$}
%     \\ \ \\
%     $^\diamond$First Affiliation \\
%     First Address 1 \\
%     First Address 2 \\
%     \texttt{first.email@example.com}
%      \\ \ \\
%      $^\dagger$Second Affiliation \\
%     Second Address 1 \\
%     Second Address 2 \\
%     \texttt{second.email@example.com}
%      \\ \ \\
%     $^\ddagger$Third Affiliation \\
%     Third Address 1 \\
%     Third Address 2 \\
%     \texttt{third.email@example.com}
% \end{quote}
  
% \fi

\end{document}